# Physics Consistency and Latent Dynamics in Spatiotemporal Physics Field Generation

Peimian Du[a], Jiabin Liu[a,*] Xiaowei Jin[a], Wangmeng Zuo[b], Hui Li[a,b,c]

[a] *Key Lab of Smart Prevention and Mitigation of Civil Engineering Disasters of the Ministry of Industry and Information Technology, Harbin Institute of Technology, Harbin, 150090, China*

[b] *School of Computer Science and Technology, Harbin Institute of Technology, Harbin, 150090, China*

[c] *Guangdong-Hong Kong-Macao Joint Laboratory for Data-Driven Fluid Mechanics and Engineering Applications, Harbin Institute of Technology (Shenzhen), Shenzhen 518055, China*

## Abstract

Data-driven models for spatiotemporal physical field generation, such as flow and acoustic fields, often deviate from governing equations and lack interpretability in latent temporal dynamics. To address these challenges, we propose HMT-PF, a hybrid Mamba-Transformer architecture for physical field generation. The framework incorporates a query-based gradient computation mechanism and a physics-informed fine-tuning strategy to enhance physical consistenc. Analysis of the latent space reveals that the initial latent state vector evolves as an autonomous dynamical system under the Mamba backbone. Principal component analysis (PCA) indicates that a small number of dominant modes in initial latent state vector govern the key evolution patterns of the physical field, while a Jacobian-based temporal sensitivity analysis characterizes the intrinsic dynamical structure and stability of the latent evolution. Experiments across five benchmark datasets demonstrate strong performance, and physics-informed fine-tuning further reduces physical residuals, highlighting the effectiveness of the proposed latent-level fusion strategy. An empirical scaling law between prediction error and physical residual is identified, revealing a consistent exponential relationship in the low-error regime. Based on this observation, an dual-metric framework is proposed to jointly evaluate numerical accuracy and physical realism.

## I. Introduction

Natural physical fields including flow fields and acoustic fields are described by partial differential equations (PDEs). The solving of such PDEs is extremely significant in a great variety of scientific and engineering fields such as aerospace, mechanical engineering, civil engineering, marine engineering as well as medicine engineering [1-5]. Traditional methods for solving physical problems through numerical simulations primarily involve resolving these

* Corresponding authors.

E-mail addresses: liujiabin@hit.edu.cn (J. Liu).

equations using numerical discretization techniques such as Finite Difference Method and Finite Volume Method [6, 7]. However, these conventional approaches can be time-consuming and costly, especially involving complex geometries. Recently, machine learning methods have shown potential in addressing PDEs encountered in science and engineering. Learning solution to PDEs has been proposed under two paradigms: 1) data-driven learning and 2) physics-informed optimization. The former utilizes data from existing solvers or experiments, while the latter is purely or partly based on PDEs constraint [8].

In data-driven training of neural network models for physical field generation, several impactful approaches can be broadly classified based on their spatial discretization methods. These approaches can be categorized into two main types: 1) discretization-dependent methods, and 2) discretization-independent methods [9]. The former approach necessitates partitioning the data domain into a specific grid or mesh. CNN is first implemented on regular grids for learning transient fluid dynamics [10]. And the TF-Net [11] aims to predict turbulent flow by learning from the highly nonlinear dynamics of spatiotemporal velocity fields. The Fourier Neural Operator (FNO) [12] marks a significant development in neural operators by innovatively parameterizing the integral kernel directly in Fourier space, thus creating an expressive and efficient architecture. Expanding on this concept, U-FNO [13], which introduce the U-Net model as an enhancement, is designed based on original FNO. Another remarkable advance in neural operator is DeepONet [14] which marks the transition towards learning mappings between functional spaces. PI-DeepONet [15] uses PDE residuals for unsupervised training, enhancing the ability to learn without explicit supervision. MIONet [16] expands on DeepONet by introducing a neural operator with branch nets for input function encoding and a trunk net for output domain encoding. DeepOKAN [17], a deep operator network based on Kolmogorov-Arnold Networks (KANs) with Gaussian radial basis functions (RBFs), demonstrating superior accuracy over traditional DeepONets in tasks such as 1D wave propagation, 2D orthotropic elasticity, and transient Poisson problems. However, the domain that discretization-dependent methods can address is limited. The latter approach does not rely on any discretization techniques which can deal with any unstructured grid. MGNO [18] utilizes a class of integral operators, with the kernel integration being computed through graph-based message passing. Geo-FNO [19] extends the FNO by learning geometric deformations, enabling fast Fourier transform on irregular domains. GINO [20] integrates graph neural networks and Fourier architectures to learn solution operators on irregular meshes. Because attention mechanism has the capability of capturing long-term dependency relationship, Transformers have achieved remarkable success in a great variety of fields [21-30], which have been introduced in PDEs solving as well. The Galerkin [31] Transformer proposes a SoftMax-free attention mechanism for PDEs operator learning, it achieves computational efficiency while maintaining strong performance on tasks like Burgers' equation and Darcy flow. OFormer [32] and UPT [33] utilizes an attention-based framework with propagating the latent space vectors for learning PDE solution operators. GNOT [34] presents a scalable and flexible transformer-based framework, leveraging a heterogeneous normalized attention layer and a geometric gating mechanism to enhance operator learning. Transolver [35] is a fast Transformer-based PDE solver that introduces Physics-Attention to decompose discretized domains into learnable slices representing physical states. A novel latent neural PDE solver named LNS [36] that decouples dynamics prediction from spatial discretization through a autoencoder-propagator framework. The models aforementioned are trained by supervised learning with input-output data pairs and provide the possible to generate the physical field without directly solving their governing equations.

Data-driven methods in model training often generate datasets that resemble ground truth data in certain aspects. However, a closer examination of the underlying physical laws reveals significant discrepancies in the governing equations, primarily due to limitations in both model architecture and training dataset size. Models based on data-driven learning face certain challenges in incorporating physical mechanisms and still rely on the data itself to satisfy specific physical laws. In an ideal scenario where both the model and dataset are infinitely large and encompass all possible situations, it is likely that the model would generate results consistent with fundamental physical laws. Nevertheless, such conditions are impractical, as current models already require substantial computational resources. Furthermore, the complexity and diversity inherent in physical phenomena make the generation of datasets a considerable engineering challenge. For a new generation test where ground truth data are unknown, the most important criterion for assessing the quality of the generated field is whether it satisfies its governing physical equations.

To improve the physical realism of model predictions, researchers try to use the physical equation constraints in the loss function. Physics-informed models, such as Physics-Informed Neural Networks (PINNs), employs a neural network to represent the solution function and minimizes a loss function to diminish the violation of the specified governing equations [37]. NSFnets [38] provides a PINNs approach for solving the incompressible Navier-Stokes equations, demonstrating capabilities in simulating both laminar and turbulent flows. PINNsFormer [39] enhances PINNs using Transformers to model temporal dependencies in PDEs, which introduces a Wavelet activation function for better approximation. The potential of PINNs in solving PDEs across various applications, including high-dimensional problems, has been demonstrated in recent work [40]. Physic GNN [41] utilizes graph neural network achieving higher accuracy than traditional PINNs, while maintaining real-time computational efficiency. Beyond soft penalties, the method [42] regards PDE constraints as hard constraints via augmented Lagrangian relaxation within a multi-fidelity data fusion framework is investigated. It is important to note, however, that PINNs are designed to represent the solution function for a specific instance, rather than mapping input functions to output solution functions. As a result, when confronted with a new task, it may be necessary to redefine the loss function and training process to accommodate new inputs and boundary conditions.

To address the generalization problem of data-driven trained models, fine-tuning has emerged as a prevalent technique in machine learning [43], involving the adaptation of a pre-trained model to a specific task or dataset. This process encompasses the retraining of the model on a smaller, task-specific dataset, thereby enhancing its performance within the context of the intended application. By leveraging the knowledge encapsulated within the pre-trained model, fine-tuning facilitates the efficient transfer of learned features, allowing the model to generalize more effectively to the unique characteristics of the new task. This methodology is particularly beneficial in scenarios where labeled data is limited, as it mitigates the computational resources and time required for training, while still achieving satisfactory performance levels. Fine-tuning methods can be categorized into Addition-based Tuning, Partial-based Tuning and Unified-based Tuning [43, 44]. These methods, such as Adapter Tuning [45], Prompt Tuning [46], and LoRA [47], vary in the way they modify model parameters or introduce new learnable components to improve efficiency and flexibility.

In this study, motivated by the complementary strengths of Mamba and Transformer architectures [25, 26, 48-57] in sequence modelling and global context extraction, we propose a hybrid framework of Mamba-Transformer for spatiotemporal physical field generation on unstructured grids. The framework unifies data-driven spatiotemporal modelling with a physics–data fusion refinement module grounded in finite difference gradient estimation, and is further accompanied by a comprehensive analysis of latent dynamics and physical interpretability of the model's internal representations.

## II. Methodology

### *A. Neural network*

Fig. 1 illustrated the architecture of the proposed hybrid Mamba-Transformer network with physics-data fusion (HMT-PF) model, including a backbone network and a fine-tuning block. The backbone network consists of three main components: (i) an encoder, $\mathcal{E}_1$, that flexibly encodes disorder input into a unified latent representation of shape; (ii) a Mamba block, $\mathcal{M}$, facilitates forward time propagation within the latent space, allowing for the progression from the initial input to the designated target time; (iii) a decoder block, $\mathcal{D}$, and a FFN transfers the latent representation into physical properties $\phi(\boldsymbol{x}_Q, t)$ at specific query positions. In addition, a physics-informed fine-tuning block, $\mathcal{T}$, computes the residuals of the physical equations, which are subsequently encoded by an encoder $\mathcal{E}_2$. These residuals encoding is used to refine the latent representation, ensuring that the generated field properties adhere to the physical laws.

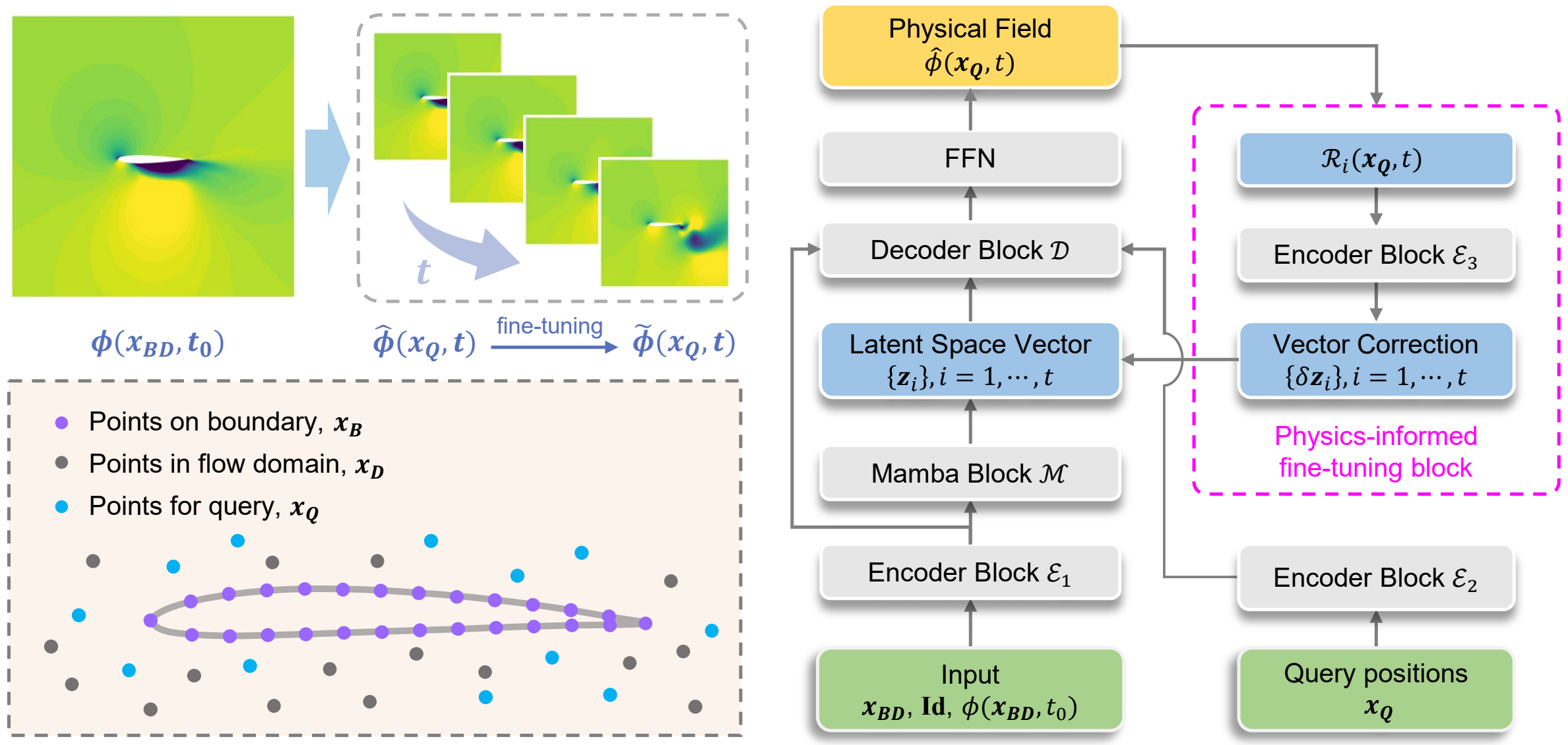

**Fig. 1.** Overview of hybrid Mamba-Transformer network with physics-data fusion (HMT-PF) model.

*B. Input information*

The input to HMT-PF primarily consists of coordinates, identifiers, and initial condition information. **Coordinates:** The input point coordinates are divided into two categories, $\boldsymbol{x_{BD}}$ and $\boldsymbol{x_Q}$, as illustrated in Fig. 1. $\boldsymbol{x_{BD}}$ is the union set of $\boldsymbol{x_B}$ and $\boldsymbol{x_D}$, which represent the coordinates corresponding to specific points on the object boundary and within the domain, respectively. Additionally, $\boldsymbol{x_Q}$ refers to the coordinates at query positions where specific physical properties will be generated by the network. They have $\boldsymbol{x_B} \in \mathbb{R}^{N_B \times d}$, $\boldsymbol{x_D} \in \mathbb{R}^{N_D \times d}$ and $\boldsymbol{x_Q} \in \mathbb{R}^{N_Q \times d}$, where $N_B$, $N_D$ and $N_Q$ are the point number of them, respectively. $d$ is the spatial dimension of the physical problem. **Identifier:** To distinguish the inputs from $\boldsymbol{x_B}$ and $\boldsymbol{x_D}$, we have introduced a unique identifier to differentiate between them. Specifically, an identity parameter **Id** is assigned, where **Id**=0 corresponds to $\boldsymbol{x_B}$, and **Id**=1 corresponds to $\boldsymbol{x_D}$. For other types of points, they can be distinguished by setting additional **Id. Initial condition:** when the initial state is given, the physical field $\phi$ at location $\boldsymbol{x_{BD}}$ are used as input.

*C. Encoder block $\mathcal{E}_1$*

Encoder block $\mathcal{E}_1$ is designed to extract the feature of input data, with its primary objective being to learn how to selectively focus on the most pertinent aspects, which is shown in Fig. 2.

Encoder block has:

$$\mathcal{G}_0 = \mathcal{E}_1(\boldsymbol{x_{BD}}, \mathbf{Id}, \phi(\boldsymbol{x_{BD}}, t_0)) \quad (1)$$

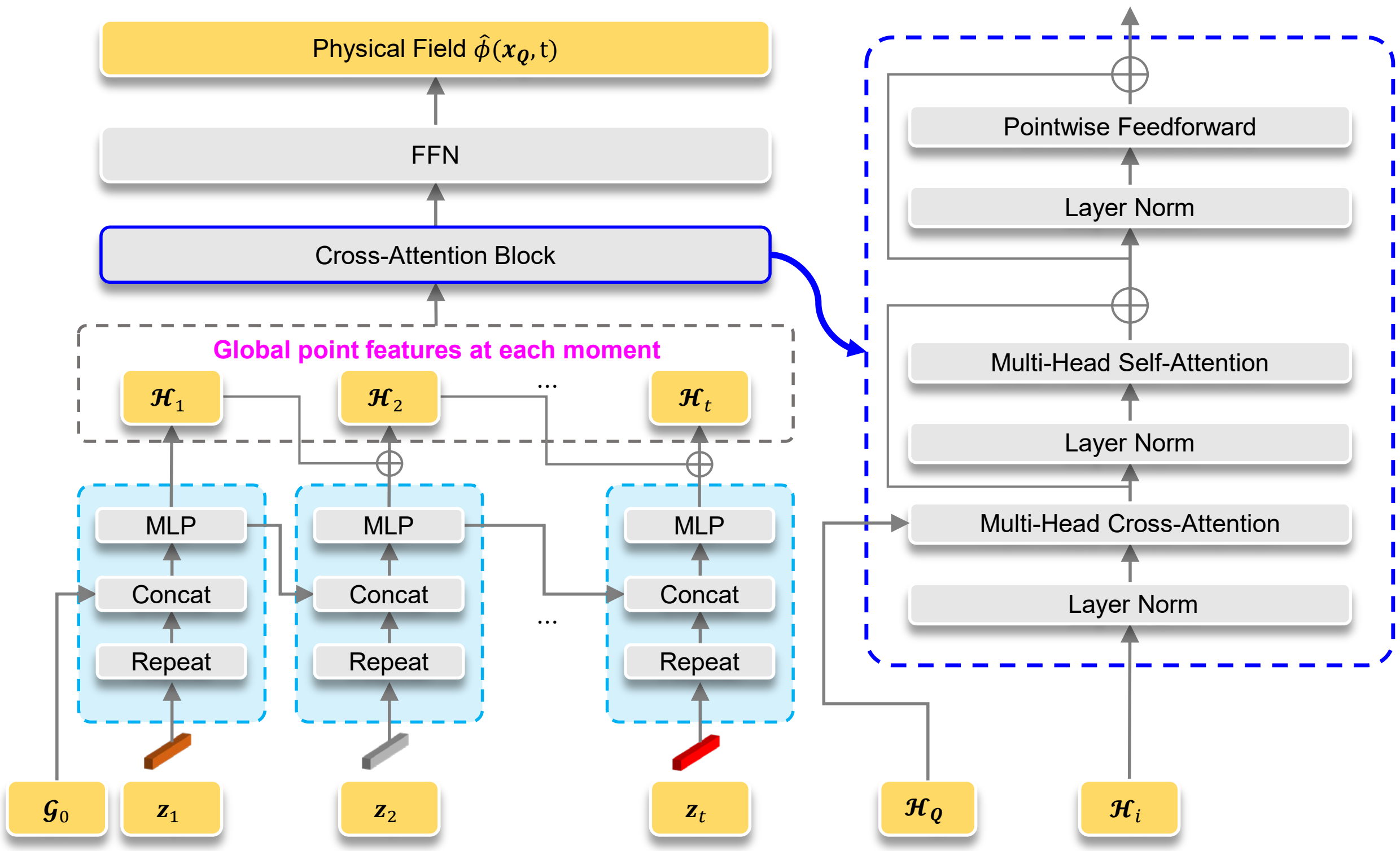

**Fig. 2.** Schematic of the decoder architecture.

where $\mathcal{G}_0 \in \mathbb{R}^{N_D \times N_g}$ is the global feature derived from the geometric input and the initialization physical field input. The input data are deal as following:

$$\boldsymbol{y}_1 = \mathrm{MLP}(\boldsymbol{x}_{\boldsymbol{BD}}), \quad \boldsymbol{y}_1 \in \mathbb{R}^{N_{BD} \times N_C} \tag{2}$$

$$\boldsymbol{y}_2 = \mathrm{MLP}(\mathbf{Id}), \quad \boldsymbol{y}_2 \in \mathbb{R}^{N_{BD} \times N_C} \tag{3}$$

$$\boldsymbol{y}_3 = \mathrm{MLP}\big(\phi(\boldsymbol{x}_{\boldsymbol{BD}}, t_0)\big), \quad \boldsymbol{y}_3 \in \mathbb{R}^{N_{BD} \times N_C} \tag{4}$$

where $N_C$ is the embedding dimension for the input of $\mathbf{Id}$. A concatenation is used in $\mathcal{E}_1$ to combine features from different data sources. After that, we use a multilayer perceptron to get the combined features called $\boldsymbol{y}_{fusion} \in \mathbb{R}^{N_{BD} \times N_g}$.

Following the extraction of the combined features, a local feature embedding block is applied to refine the feature representation. For each grid point $\boldsymbol{x} \in \boldsymbol{x}_{\boldsymbol{BD}}$, the set of $k$-nearest neighbors is denoted as $\text{KNN-Grouping}_k(\boldsymbol{x}, \boldsymbol{x}_{\boldsymbol{BD}})$, which represents a local subset of the grid points within $\boldsymbol{x}_{\boldsymbol{BD}}$. The $k$-nearest neightbor (KNN) algorithm is widely used for capturing local geometric structures in point clouds and spatial data [59, 60]. Based on the KNN-Grouping, the local feature embedding block generates the output feature [61] $\boldsymbol{y}_l \in \mathbb{R}^{N_{BD} \times N_g}$, computed by:

$$\Delta\boldsymbol{y}_{fusion}(\boldsymbol{x}) = \mathrm{concat}_{\boldsymbol{x}_k \in \text{KNN-Grouping}_k(\boldsymbol{x}, \boldsymbol{x}_{\boldsymbol{BD}})}\big(\boldsymbol{y}_{fusion}(\boldsymbol{x}_k) - \boldsymbol{y}_{fusion}(\boldsymbol{x})\big) \tag{5}$$

$$\boldsymbol{y}_l(\boldsymbol{x}) = \mathrm{Maxpooling}(\mathrm{MLP}(\mathrm{concat}(\Delta\boldsymbol{y}_{fusion}(\boldsymbol{x}), \mathrm{Repeat}(\boldsymbol{y}_{fusion}(\boldsymbol{x}))))) \tag{6}$$

where $\boldsymbol{y}_{fusion}(\boldsymbol{x})$ represents the fused feature for grid point $\boldsymbol{x}$.

The resulting feature $\boldsymbol{y}_l$ is subsequently processed through Galerkin self-attention [31] to generate a more comprehensive representation, $\mathcal{G}_0$. The computation process for Encoder block $\mathcal{E}_1$ proceeds as follows:

$$\{\boldsymbol{y}_1, \boldsymbol{y}_2, \boldsymbol{y}_3\} \xrightarrow{\text{concat and MLP}_{fusion}} \boldsymbol{y}_{fusion} \in \mathbb{R}^{N_{BD} \times N_g} \xrightarrow{\substack{\text{local feature} \\ \text{embedding}}} \boldsymbol{y}_l \in \mathbb{R}^{N_{BD} \times N_g} \xrightarrow{\text{self-attention}} \mathcal{G}_0 \in \mathbb{R}^{N_{BD} \times N_g} \tag{7}$$

*D. Mamba block* $\mathcal{M}$

In this block, a max pooling layer is applied to extract the primary features from $\mathcal{G}_0$, which are then represented by $\boldsymbol{z}_0$:

$$\boldsymbol{z}_0 = \text{Maxpooling}(\mathcal{G}_0), \ \boldsymbol{z}_0 \in \mathbb{R}^{1 \times N_g} \tag{8}$$

where $\boldsymbol{z}_0$ is the global aggregation feature at the given initial moment, and need to be propagated in the latent space over time to capture its evolution. To achieve this, a latent space dynamic propagation module based on Mamba module is introduced. The Mamba module [58] can effectively capture long-range dependencies in time series data, making it suitable for dynamic modeling. The global aggregation feature at the initial moment generates subsequent time step features through the autoregressive mechanism of the Mamba module. Then, the global aggregation feature $\boldsymbol{z}_i$ at time step $i$ is obtained by:

$$\boldsymbol{z}_i = Mamba(\boldsymbol{z}_0, \boldsymbol{z}_1, \cdots, \boldsymbol{z}_{i-1}) \tag{9}$$

This autoregressive generation mechanism ensures that the global aggregation features $\boldsymbol{z}_i$ can integrate all historical information, thereby capturing the spatiotemporal evolution in complex dynamic scenarios. The introduction of the Mamba module enhances the network's temporal modeling capabilities, particularly in capturing potential nonlinear relationships and dynamic dependencies across continuous time steps.

*E. Encoder block* $\mathcal{E}_2$

To conduct point queries within the latent space, it is need to map the query point coordinates into this space. By using the multi-layer perceptron, this mapping can be effectively performed:

$$\mathcal{H}_Q = \text{MLP}(\boldsymbol{x}_Q), \ \mathcal{H}_Q \in \mathbb{R}^{N_Q \times N_g} \tag{10}$$

in which the input is the coordinates $\boldsymbol{x}_Q$ of query position and $\mathcal{H}_Q$ is the latent vector after the encoding.

*F. Decoder block* $\mathcal{D}$

As shown in Fig. 2, after obtaining the global aggregation features $\boldsymbol{z}_i$ for each time step, the next step involves feature fusion, combining the initial global point features $\mathcal{G}_0$ with the global aggregation features at each time step to generate the latent space global point features $\{\mathcal{H}_i \in \mathbb{R}^{N_z \times N_g}\}_{i=1}^{t}$ for each time step. This process leverages point-wise processing of the point set and the feature fusion mechanism of the multilayer perceptron, recursively propagating and updating the global point features in the latent space. Specifically, the initial global point features $\mathcal{G}_0$ is fused with the global aggregation feature $\boldsymbol{z}_1$ at time step 1 to obtain the global point features $\boldsymbol{\mathcal{H}}_1 \in \mathbb{R}^{N_z \times N_g}$, then the global point features at time step 1 are fused with the global aggregation features at time step 2 to obtain the global point features at time step 2 $\boldsymbol{\mathcal{H}}_2 \in \mathbb{R}^{N_z \times N_g}$, and so on. Here, the MLP is invariant to the permutation of the point set, meaning that regardless the input order of the point set, the output remains unchanged, only the order of the output point set is adjusted. Therefore, this module can utilize any network that satisfies the permutation invariance of the point set.

After obtaining the global point feature vector, $\boldsymbol{\mathcal{H}}_i$, a Block based on Galerkin Cross-Attention [31] with shared weights is used for generate $\hat{\phi}(\boldsymbol{x_Q}, t)$ at query points. It computes multi-head cross-attention between the high-dimensional features of query points and the global point features at each time step, integrating the positional information with the field information at current time step. Additionally, it maintains the permutation invariance of the query point set during the information fusion process. Since each query point performs cross-attention computation with the latent space global point features separately, the cross-attention mechanism also allows for an arbitrary number of query points without affecting the high-dimensional feature vectors obtained for each query point. Finally, a weight-sharing FFN is used to derive the corresponding flow field information for each query point. The output of decoder block $\mathcal{D}$, denoted as $\hat{\phi}(\boldsymbol{x}_Q, t)$**,** is computed by the following equation:

$$\hat{\phi}(\boldsymbol{x}_Q, t) = \text{FFN} \circ \mathcal{D}\left(\boldsymbol{z}_1, \cdots, \boldsymbol{z}_t; \mathcal{G}_0, \boldsymbol{\mathcal{H}}_{\boldsymbol{Q}}\right) \tag{11}$$

The parameters of the FFN and the Cross-Attention block are shared across all time steps. This design not only reduces the number of model parameters but also ensures smooth propagation of features between different time steps, improving training efficiency and generalization ability of the network.

*G. Physics-Data fusion fine-tuning block* $\mathcal{T}$

The ground truth values of $\phi$ are expected to be obtained from field measurements, experiments, or numerical simulations, since they already satisfy the governing physical equations. However, during training, the model was not explicitly trained on the gradients of predicted physical properties or second-order derivatives. Due to the physical equations are sensitivity to gradient terms, these errors remain significant. A Physics-Data fusion fine-tuning process is introduced for each sample based on physical information to alleviate this issue. Leveraging the flexible query mechanism, spatial derivatives at a query point $x_Q$ are computed using finite difference method [62, 63] by sampling neighboring points spaced at intervals $\Delta x_i$:

$$\left.\frac{\partial \varphi}{\partial x_i}\right|_{x_Q} = \frac{\varphi(x_Q+\Delta x_i, t) - \varphi(x_Q-\Delta x_i, t)}{2\Delta x_i} \tag{12}$$

$$\left.\frac{\partial^2 \varphi}{\partial x_i^2}\right|_{x_Q} = \frac{\varphi(x_Q+\Delta x_i, t) - 2\varphi(x_Q, t) + \varphi(x_Q-\Delta x_i, t)}{(\Delta x_i)^2} \tag{13}$$

Similarly, the temporal derivative is computed by:

$$\left.\frac{\partial\varphi}{\partial t}\right|_{x_Q} = \frac{\varphi(x_Q,t+\Delta t)-\varphi(x_Q,t)}{\Delta t} \tag{14}$$

Taking transient flow of Airfoil dataset as an example, the physical properties $(\boldsymbol{u}, p, \rho)$ is predicted using the neural network of the data-driven module, where $\boldsymbol{u}$ is the velocity vector, $p$ is the pressure and $\rho$ is the density. The N-S equation residuals [64, 65] at each spatial point and time step is computed and their residuals have

$$\mathcal{R}_1(\boldsymbol{x}_Q, t) = \partial_t\hat{\rho} + \nabla\cdot(\hat{\rho}\hat{\boldsymbol{u}}) \tag{15}$$

$$\boldsymbol{\mathcal{R}}_2(\boldsymbol{x}_Q, t) = \partial_t(\hat{\rho}\hat{\boldsymbol{u}}) + \nabla\cdot(\hat{\rho}\hat{\boldsymbol{u}}\otimes\hat{\boldsymbol{u}}) + \nabla\hat{p} \tag{16}$$

in which $\mathcal{R}_1(\boldsymbol{x}_Q, t)$ is the residual of the continuity equation; $\boldsymbol{\mathcal{R}}_2(\boldsymbol{x}_Q, t)$ is the residual of the momentum equation in different direction.

Fig. 3(a) illustrates of the forward process in the Physics-Data fusion fine-tuning block. One of the key steps of this block is the feature encoding of the residual matrix $\mathcal{R}_1$ and $\boldsymbol{\mathcal{R}}_2$. To capture the spatial and temporal correlation of the residual, encoder $\mathcal{E}_3$ extracts the correction term $\delta\boldsymbol{z} \in \mathbb{R}^{T\times d}$ from the residual matrix:

$$\delta\boldsymbol{z} = \mathcal{E}_3(\mathcal{R}_1, \boldsymbol{\mathcal{R}}_2) = \{\delta\boldsymbol{z}_i\}_{i=1}^{T} \tag{17}$$

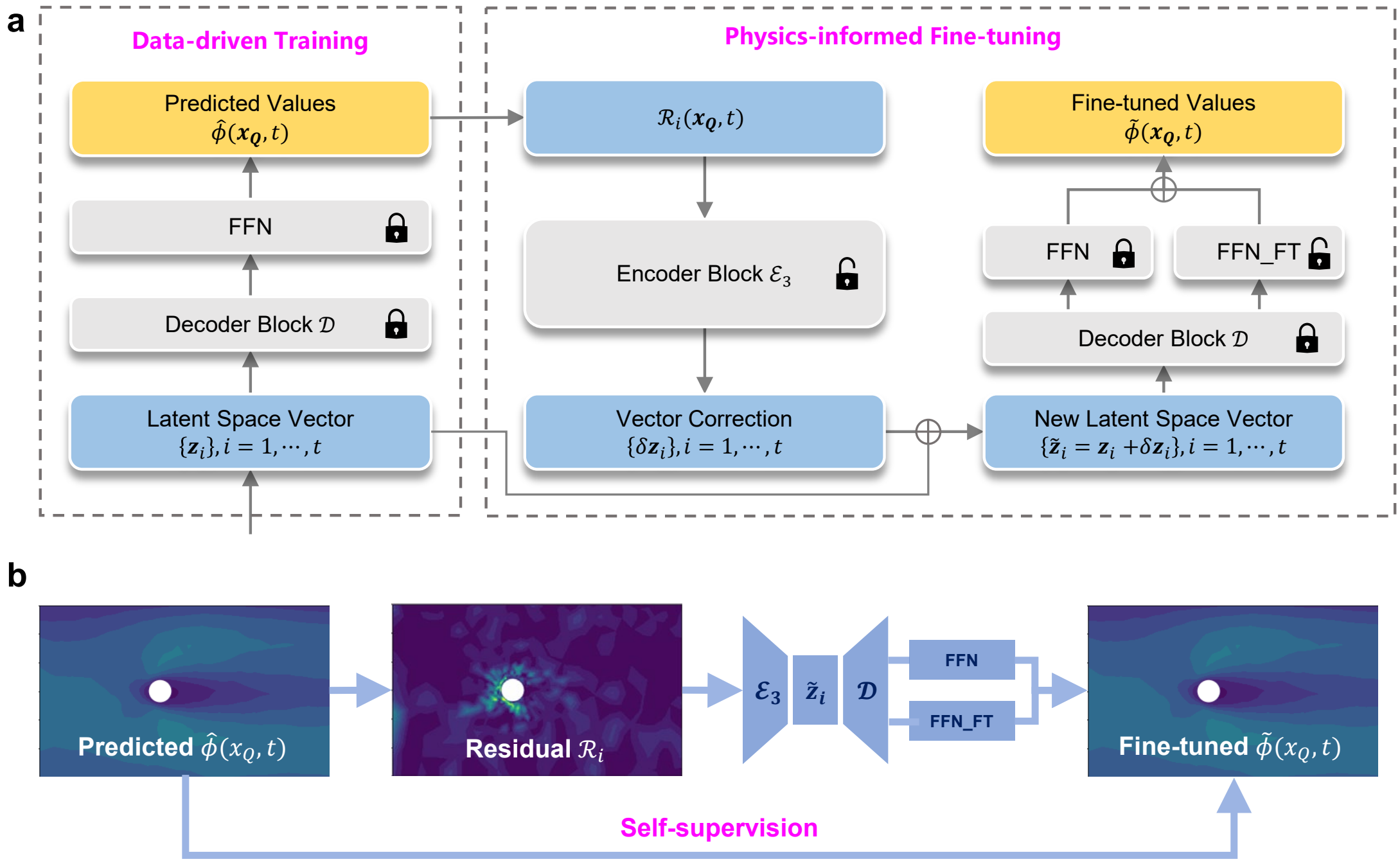

**Fig. 3.** Forward process in the fine-tuning block with Physics-Data Fusion Fine-tuning framework. **a.** Fine-tuning framework. **b.** Self-supervision term.

Physics-Data fusion fine-tuning module combines the correction term with the original latent features to obtain the latent fused space features $\tilde{\boldsymbol{z}} = \{\delta\tilde{\boldsymbol{z}}_i\}_{i=1}^{T} \in \mathbb{R}^{T\times d}$, where $\tilde{\boldsymbol{z}} = \boldsymbol{z} + \delta\boldsymbol{z}$. To further integrate the features and predict the temporal flow field, FFN_FT decodes the latent fused space feature vector to produce the final temporal flow field prediction:

$$\tilde{\phi}(\boldsymbol{x}_Q, t) = \text{FFN} \circ \mathcal{D}(\tilde{\boldsymbol{z}}; \boldsymbol{\mathcal{G}}_0, \boldsymbol{\mathcal{H}}_{\boldsymbol{Q}}) + \text{FFN_FT} \circ \mathcal{D}(\tilde{\boldsymbol{z}}; \boldsymbol{\mathcal{G}}_0, \boldsymbol{\mathcal{H}}_{\boldsymbol{Q}}) \tag{18}$$

where $\mathcal{D}(\tilde{\boldsymbol{z}}; \boldsymbol{\mathcal{G}}_0, \boldsymbol{\mathcal{H}}_{\boldsymbol{Q}})$ denotes the decoding operation applied to the corrected latent space features; FFN and FFN_FT are feedforward networks used for feature fusion and temporal flow field prediction; $\tilde{\phi}(\boldsymbol{x}_Q, t)$ is physical field prediction after fine-tuning, aligns more closely with physical laws.

In the Physics-Data fusion fine-tuning module, only the parameters of the residual feature encoder $\mathcal{E}_3$ and the feedforward network FFN_FT are trainable, while the parameters of the other modules are fixed, as shown in Fig. 3. In a Physics–Data fusion manner, the residuals of the physical equations in the predictions obtained from data-driven training are first calculated. As shown in Fig. 3(a), the encoded residual features are then fused into the latent space vector to update. The fine-tuned prediction is subsequently obtained through an additional decoding step. Fig. 3(b) demonstrates the process of preventing the physical field from deviating significantly after fine-tuning, only a subset of grid points is constrained, ensuring stable and physically consistent updates. This design ensures that during the fine-tuning process, only the parts related to the physical constraints are updated, thus preventing unnecessary modifications to the data-driven model itself. Simultaneously, freezing the parameters of the other modules reduces the computational load during the fine-tuning process, significantly improving the efficiency of both the training and inference stages.

## III. Model train

The training of the model consists of two stages. Initially, the parameters within the encoder blocks $\mathcal{E}_1$ and $\mathcal{E}_2$, Mamba block $\mathcal{M}$, and the decoder block $\mathcal{D}$ are trained using the input-output data pairs. Subsequently, a fine-tuning mechanism is employed, where no additional data is required for training.

### *A. First Stage: Data-Driven training*

In the first training step, the input data is encoded into latent representations, processed by the dynamic propagation module, and then decoded to reconstruct the output. The training is guided by the loss function:

$$\mathcal{L}_1 = \sum_{i=1}^{N_\phi} \frac{1}{N_Q T} \left\| \hat{\phi}_i - \phi_i \right\|_2^2 \tag{19}$$

where $\hat{\phi}_i$ denotes the $i$-th physical property of predicted outputs, and $\phi_i$ represents its corresponding ground truth (GT) value. $T$ refers to the number of time steps, $N_Q$ is the number of query points, and $N_\phi$ denotes the number of physical properties. Gradient-based optimization minimizes $\mathcal{L}_1$ to update the model parameters, ensuring accurate reconstruction and effective learning of the latent space dynamics.

### *B. Second Stage: Physics-informed fine-tuning*

During training of fine-tuning, the loss function during the fine-tuning phase consists of two components: the data self-supervision term and the physical conservation term, as shown in Fig. 3(b). Based on this, the overall loss function during the fine-tuning phase is formulated as:

$$\mathcal{L}_2 = \lambda_\phi \sum_{i=1}^{N_\phi} \frac{1}{N_Q T \xi_i} \left( \boldsymbol{M}_i \odot \left\| \tilde{\phi}_i - \hat{\phi}_i \right\|_2^2 \right) + \lambda_{\mathcal{R}i} \sum_{i=1}^{N_R} \frac{1}{N_Q T} \left\| \tilde{\mathcal{R}}_i \right\|_2^2 \tag{20}$$

where $\xi_i$ is the propotion of the data used for self-supervision training. $\boldsymbol{M}_i$ is random mask matrixes using for random sampling data of $\phi_i$. $\odot$ represents the element-wise multiplication (Hadamard product). $\tilde{\mathcal{R}}_i$ represents the equation residuals computed at each querying point using the updated latent space feature vector; $\lambda_\phi$ and $\lambda_{\mathcal{R}i}$ are hyperparameters used to balance the weights between the self-supervision term and the physical conservation term. Since ground truth values are not employed in this process, the loss function is constructed using the predicted values with and without fine-tuning. This constitutes a self-supervised training procedure. Consequently, during the data-driven training phase, the generated physical field need to approximate the ground truth values as closely as possible. The fine-tuning process serves to refine the original predictions, thereby ensuring that the results more effectively adhere to the governing physical equations.

## IV. Experiments

### *A. Dataset*

We explore a range of models on various datasets, including the Airfoil dataset, Simple car dataset, Aneurysm dataset, and Acoustic dataset, shown as Table 1. Except for the simple car datasets, all datasets are time-dependent. For the simple car datasets, the objective is to predict the average pressure and average stress in three directions based on the shape of the car as input. For the remaining datasets, the goal is to predict a series of physical fields, such as velocities and density over time, based on the initial conditions of them. The details of these datasets is written below.

**Table 1.** Overview of evaluated datasets.

| Data type | Dataset name | # Dim | # Point number |
|---|---|---|---|
| Point cloud | Airfoil [66] | 2D + Time | 5,233 |
| | Cylinder [66] | 2D + Time | 1,732 ~ 2,059 |
| | Aneurysm [67] | 3D + Time | 5,000 |
| | Simple car | 3D | 10,000 |
| Regular grid | Acoustic [68] | 2D + Time | 64*64 |

### *B. Baselines*

For comparative analysis, several models were evaluated alongside our proposed model, including traditional neural operators such as FNO [12], GEO-FNO [19], and GINO [20], as well as the state-of-the-art Transolver [35]. All models, including ours, were trained using the AdamW optimizer.

For our model, we configured the number of self-attention layers to 2/4/8, the hidden feature channels to 32/64/128, and the number of Mamba layers to 2/4/8. For FNO, we employed a hidden dimension of 32/64/128 and 8/16/24 Fourier modes per dimension, with the number of Fourier layers set to 4. In the case of GeoFNO, a hidden dimension of 32/64/128 was utilized, along with 8/16/24 Fourier modes per dimension for grid resolutions of $8^3/16^3/24^3$, and the number of Fourier layers was set to 5. For GINO, grid resolutions of $16^3/32^3/48^3$ were established with 8/16/24 modes per dimension, and a message aggregation radius of 0.05/0.1 was applied, with the hidden layer dimension set to 32/64/128 and 5 Fourier layers. For Transolver, the number of Transformer layers was set to 4/8, the hidden layer dimension to 32/64/128, and the number of slices to 32/34/128. All experiments were conducted on a single NVIDIA A100 GPU. The normalized mean squared error (MSE) was computed as the evaluation metric, and the best results for each dataset were selected for comparison.

*C. Without Physics-Data fusion fine-tuning*

In the absence of Physics-Data fusion fine-tuning, experiments were conducted across all aforementioned datasets with the backbone network. The comprehensive results are summarized in the Table 2. It presents the summed MSE (Mean Squared Error) results between the predicted and ground truth (GT) physical fields (e.g., velocity, pressure, density). Details regarding the datasets and training methodologies presented in Table 2 are provided in the Appendix. Form the comparison, our model demonstrates competitive performance across all datasets, achieving the lowest error rates in four out of the five test cases: the airfoil, cylinder, aneurysm, and acoustic datasets. For the simple car dataset, our model's performance (0.0652) is comparable to Transolver's result (0.0620), with both significantly outperforming other methods. These findings indicate that our approach maintains robust accuracy across diverse physical systems while exhibiting particular strength in spatiotemporal field generation tasks. Furthermore, Fig. 4 presents a comprehensive visual comparison between our model's predictions and ground truth values across multiple datasets, along with their corresponding error distributions. In Fig. 4(a), the Airfoil dataset illustrates the temporal evolution of velocity in the $x$-direction over 10 time steps is illustrated. The first row displays the predicted velocity field, the second row presents the ground truth, and the third row highlights the error between the predicted and ground truth values. Fig. 4(b) presents the Acoustic dataset, demonstrating the temporal evolution of acoustic wave propagation from time step 1 to 9 is demonstrated. The first row exhibits the predicted acoustic density at each time step, the second row presents the ground truth, and the third row visualizes the error. In Fig. 4(c), the Simple car dataset includes three distinct flow-related fields are depicted. The prediction of the stress vector field, stress scalar field, and pressure field are shown in the first column. The second column provides the corresponding ground truth for each field, while the third column quantifies the error between the predicted and actual values. Fig. 4(d) shows Aneurysm dataset, where the velocity vector prediction at time step 4 is shown in the first image, while the second image presents the ground truth, and the third image illustrates the prediction error. Finally, Fig. 4(e) presents the Cylinder dataset, showing the velocity in the $x$-direction at time step 1 and 10 are shown. The first row displays the predictions, the second row shows the ground truth, and the third row represents the error.

**Table 2.** Comparison of MSE results across different datasets.

| Model | Airfoil | Cylinder | Simple Car | Aneurysm | Acoustic |
|---|---|---|---|---|---|
| GEO-FNO | 0.6361 | 0.0844 | 0.1583 | 0.02115 | 0.8573 |
| GINO | 0.7100 | 0.0882 | 0.0898 | 0.00091 | 0.9707 |
| TRANSOLVER | 0.4841 | 0.0358 | **0.0620** | 0.00019 | 0.4302 |
| Our | **0.4432** | **0.0260** | 0.0652 | **0.00015** | **0.4123** |

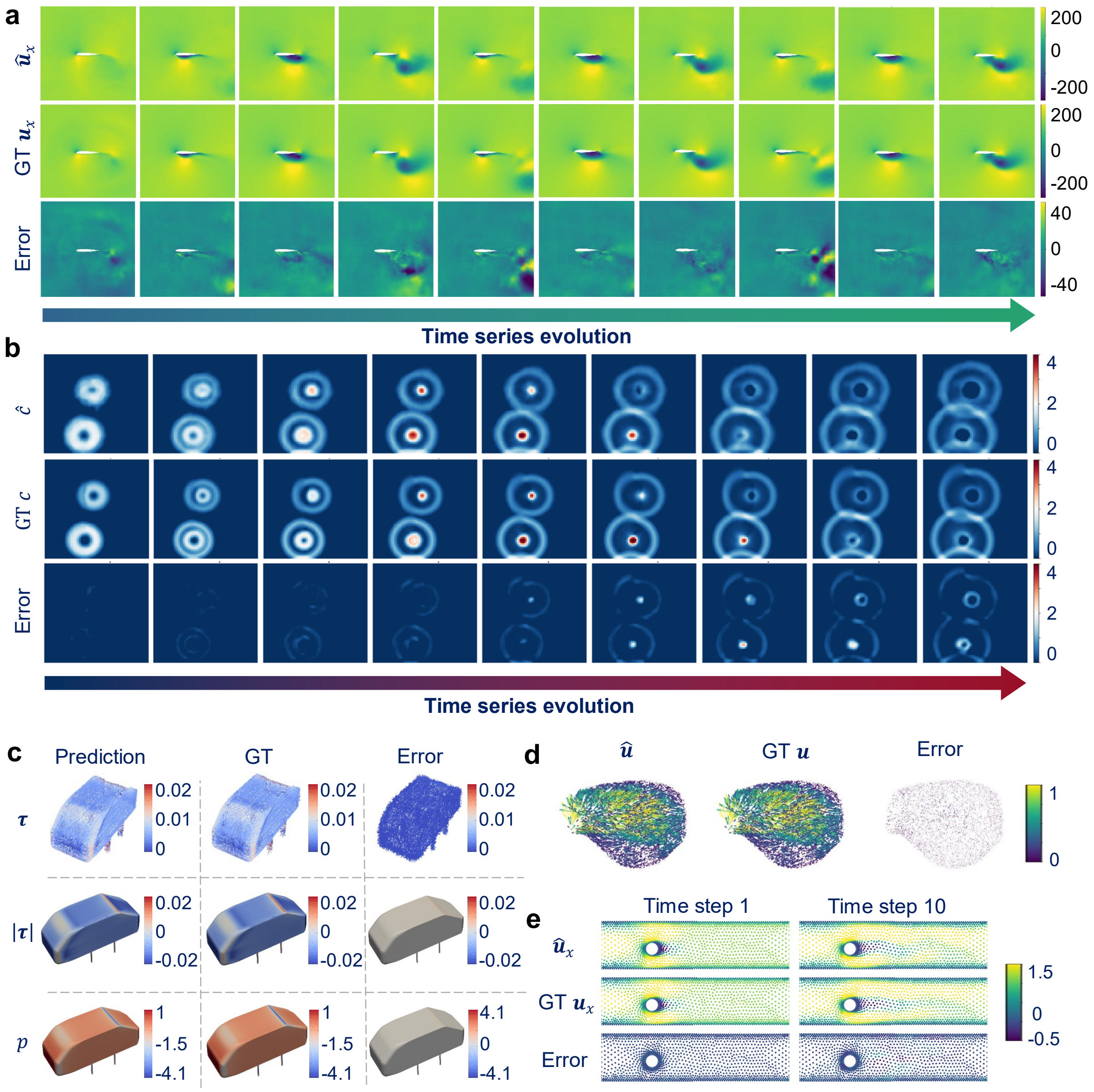

**Fig. 4.** Results display for spatiotemporal evolution across different physical phenomena. **a** Airfoil dataset. **b** Acoustic dataset. **c** Simple car dataset. **d** Aneurysm dataset. **e** Cylinder dataset.

### *D. Latent Dynamics in Mamba*

Since the temporal evolution of $\boldsymbol{z} = \{\boldsymbol{z}_i | i = 1,2,\cdots,T\}$ depends solely on $\boldsymbol{z}_0$ and the Mamba block, it can be inferred that $\boldsymbol{z}_0$ implicitly contains crucial information about the subsequent flow field evolution. To investigate the structural patterns embedded in the latent representations, we perform principal component analysis (PCA) [69, 70] on initial global aggregation vector $\boldsymbol{z}_0 \in \mathbb{R}^{128\times1}$ of airfoil dataset. As shown in Fig. 5(a), both the individual and cumulative contribution ratios of the top ten dominant principal components are computed and visualized. Remarkably, the cumulative contribution ratios of the first three principal components reaches 93.45%, indicating that the majority of variance is effectively captured. Then, $\boldsymbol{z}_0$ can be reduced into a three-dimensional space $(p^1, p^2, p^3)$ while preserving the dominant features. Clustering is subsequently performed on the latent vectors of all dataset samples across subsequent time steps $\boldsymbol{z} = \{\boldsymbol{z}_i | i = 1,2,\cdots,T\} \in \mathbb{R}^{128\times T}$ using the K-means algorithm [71, 72], as shown in Fig. 5(b). The optimal number of clusters is determined to be eight based on the silhouette score [73, 74], which quantitatively evaluates the intra-cluster compactness and separability.

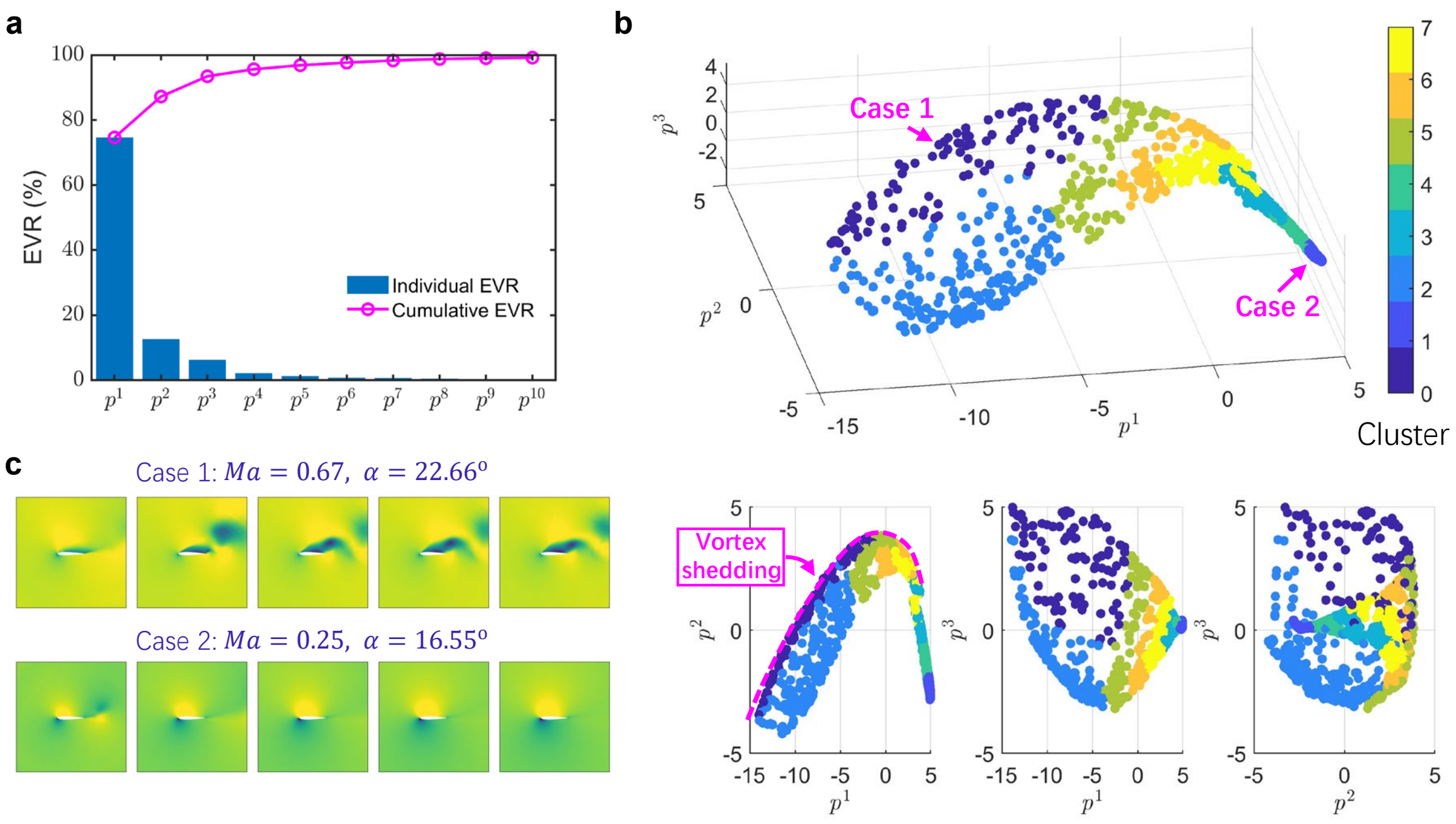

**Fig. 5.** Representative Flow Patterns from Latent Clusters and Dimensionality reduction of Latent Space via PCA. **a** Individual and cumulative contribution ratios of the five dominant eigenvalues. EVR stands for Explained Variance Ratio. $p^i$ denotes the $i^{th}$ principal component. **b** Clustering of latent vectors $\{\boldsymbol{z}_i | i = 1,2,\cdots,T\}$ projected into a three-dimensional space $(p^1, p^2, p^3)$ of $\boldsymbol{z}_0$ via PCA, with color-coded cluster assignments. **c** Selected examples with and without vortex shedding.

Through analysis, it is found that the first principal component $p^1$ is closely related to the Mach number, with a decrease in $p^1$ corresponding to an increase in the incoming Mach number $(Ma)$. As the $Ma$ increases, the flow behavior around the airfoil becomes more sensitive to variations in angle of attack $(\alpha)$ within the same range. This explains why a decrease in $p^1$ (i.e., an increase in $Ma$) leads to a progressively more discrete spatial distribution of data points. The second principal component $p^2$ shows a positive correlation with the angle of attack. Under the same $p^1$ conditions, an increase in $p^2$ typically corresponds to a larger $\alpha$ flow conditions. Therefore, flow cases with vortex shedding are primarily concentrated along the dashed boundary in the $(p^1, p^2)$ space. The third principal component $p^3$ is associated with unsteady flow characteristics. A larger value of $p^3$ indicates stronger temporal variations in the flow field. To gain physical insight into the flow characteristics represented by different clusters, two representative samples are selected, and the temporal evolution of their corresponding flow fields is visualized, as shown in Fig. 5(c).

To investigate the latent-space propagation characteristics of the Mamba block, we analyze the Jacobian matrix of the latent state at time step $t$ with respect to previous latent states at time step $\{0, \dots, t-1\}$, thereby characterising the correlation and sensitivity between the current latent output and historical latent inputs. Taking the Airfoil flow dataset as a case study, two representative flow regimes are considered for comparative analysis: a steady flow sample without vortex shedding and an unsteady flow sample with vortex shedding, as illustrated in Fig. 6(a) and Fig. 7(a), respectively.

The results indicate that, due to the time-convolution operation introduced within the Mamba block, the latent state at time step $t$ is primarily correlated with approximately the four most recent latent states at time step $\{t-4, t-3, t-2, t-1\}$, and exhibits pronounced sensitivity to the latent states at the first two preceding time steps $\{0,1\}$, as shown in Fig. 6(b,c) and Fig. 7(b,c). This suggests that, in the autoregressive generation process of latent representations, the latent states at the first two time steps play a dominant role in governing the subsequent evolution of the flow field and largely determine the fundamental pattern of flow development. As temporal recursion proceeds, the influence of latent states at intermediate time steps gradually diminishes. Consequently, the temporal propagation governed by the Mamba module is primarily dominated by the latent states at current few time steps, along with the initial two latent states. A comparative analysis of the row-wise and column-wise of the Jacobian matrix reveals distinct structural evolution patterns across different flow regimes. In the non–vortex-shedding regime, the matrix structure progressively converges toward a consistent and coherent pattern during subsequent evolution. By contrast, in vortex-shedding regime, although partial structural similarities can still persist, the overall evolution exhibits pronounced irregularity and structural disorder, as shown in Fig. 6(d,e) and Fig. 7(d,e).

The mamba block composed of ten stacked Mamba modules is employed in the network. We analyze the Jacobian matrices of the latent state at each layer with respect to the preceding latent state under non vortex shedding condition and vortex shedding condition, as shown in Fig. 8(a) and Fig. 8(b) respectively. The layer-wise Jacobian distributions exhibit highly consistent temporal correlation patterns.The results indicates that both the shallow layers and the output layer exhibit pronounced sensitivity to the preceding latent state. Notably, this sensitivity is primarily concentrated within the first two time steps and their adjacent time steps, reflecting the strong dependence of latent-state construction and output mapping on local temporal information. In contrast, from the second to the ninth layer, the sensitivity of the latent state vectors to the preceding latent state progressively diminishes with increasing depth,

becoming

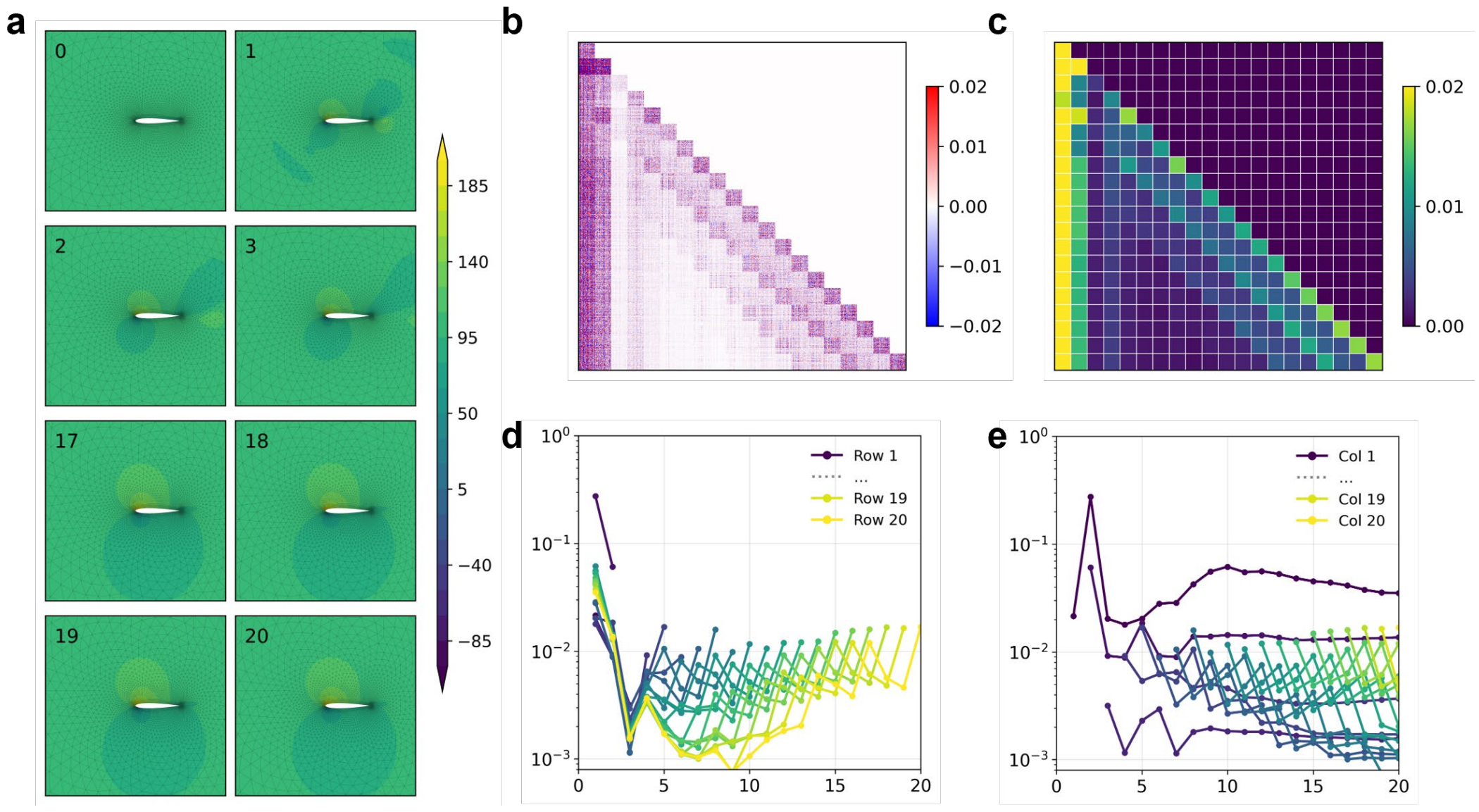

**Fig. 6.** Jacobian matrix statistics under no vortex shedding condition. **a** Airfiol flow under non vortex shedding condition. **b** Jacobian matrix of the latent statewith respect to the preceding latent state. **c** Standard deviation (std) of the Jacobian matrix elements. **d** Distribution of row-wise std values. **e** Distribution of column-wise std values.

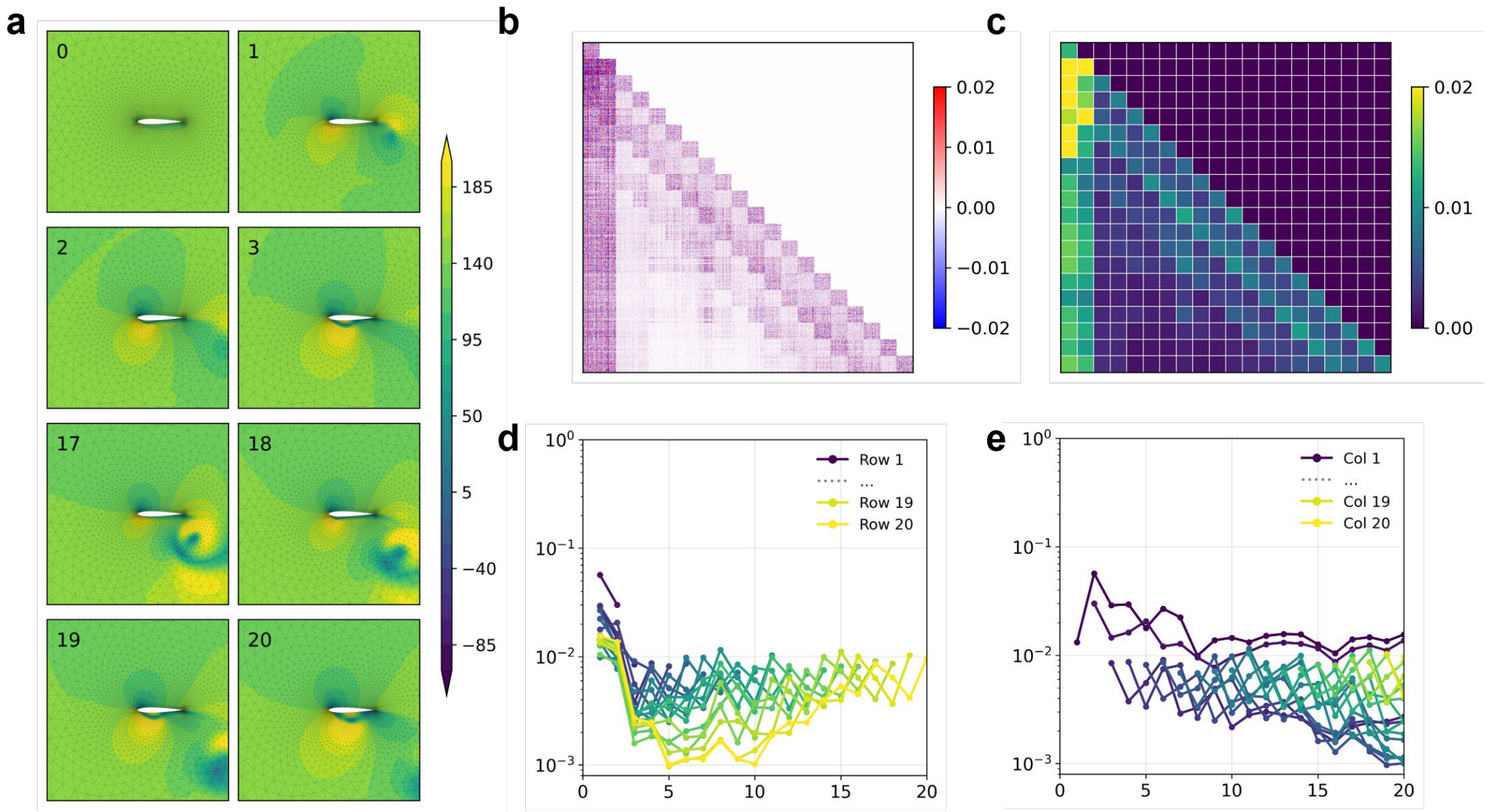

**Fig. 7.** Jacobian matrix statistics under the vortex shedding condition. **a** Airfiol flow under vortex shedding condition. **b** Jacobian matrix of the latent state with respect to thepreceding latent state. **c** Standard deviation (std) of the Jacobian matrix elements. **d** Distribution of row-wise std values. **e** Distribution of column-wise std values.

nearly negligible by the ninth layer. This behaviour suggests that layers from one to nine collectively perform an increasingly stronger encoding process, whereby latent representations are refined to enhance the network's capability to characterise the flow field while suppressing direct dependence on the initial perturbation. Interestingly, in the deepest (tenth) layer, a pronounced correlation between the latent state and the preceding latent state re-emerges. This behaviour can be interpreted as a decoding process, in which the deeply encoded latent representations are transformed to reconstruct information relevant to the output space. Overall, the stacked Mamba modules exhibit a hierarchical encode-decode mechanism during latent-state propagation.

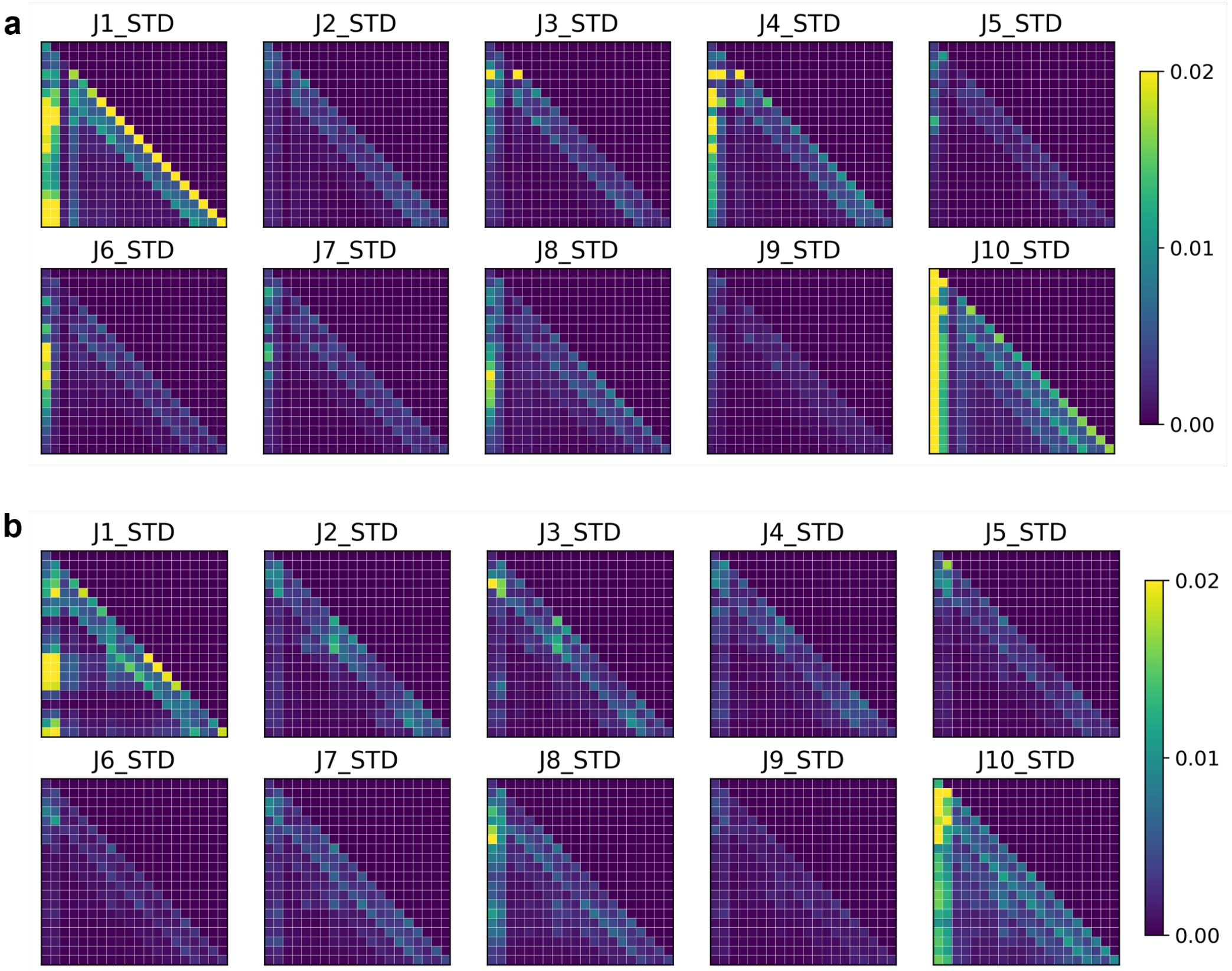

**Fig. 8.** Standard deviation of Jacobian matrices of the latent stateat different Mamba module layers. **a** Airfiol flow under non vortex shedding condition. **b** Airfiol flow under vortex shedding condition.

The latent dynamics of the entire Mamba block are primarily regulated by the final layer, which governs the evolution of latent vectors through state transitions

$$h_t = e^{\Delta_t A} h_{t-1} + \Delta_t B_t \tilde{x}_t \tag{21}$$

$$y_t = (C_t h_t + D\tilde{x}_t) \odot \mathrm{SiLU}(g_t) \tag{22}$$

$$z_t = W y_t \tag{23}$$

where $\Delta_t$, $B_t$, $C_t$, $g_t$ are all deterministic functions of the input $\tilde{x}_t$. Figure 9 provides a detailed analysis of the learned dynamics within the Mamba modules. Specifically, Fig. 9(a) shows the average value of the state matrix $A$ across the 10-layer Mamab architecture, indicating that its overall mean remains largely consistent during training. Fig. 9(b) presents the mean-centered $A$ for each layer, where inter-layer differences are primarily manifested in high-frequency components with noticeable irregularity. This suggests that variations across layers are dominated by local oscillatory features rather than global magnitude or trend. The temporal behavior of the time-step parameter $\Delta$, which is non-negative, displayed significant dynamic characteristics over time under non-vortex shedding and vortex shedding condition, as shown in Fig. 9(c) and Fig. 9(d) respectively. These subfigures further depict the state transition components $\Delta$, $e^{\Delta A}$ and $\Delta B\tilde{x}$ in the 10th Mamba layer. From a dynamical mechanism perspective, the transition matrix $e^{\Delta A}$ (where $A < 0$) is inherently a strict dissipative operator, thereby precluding spontaneous emergence of periodic oscillations.

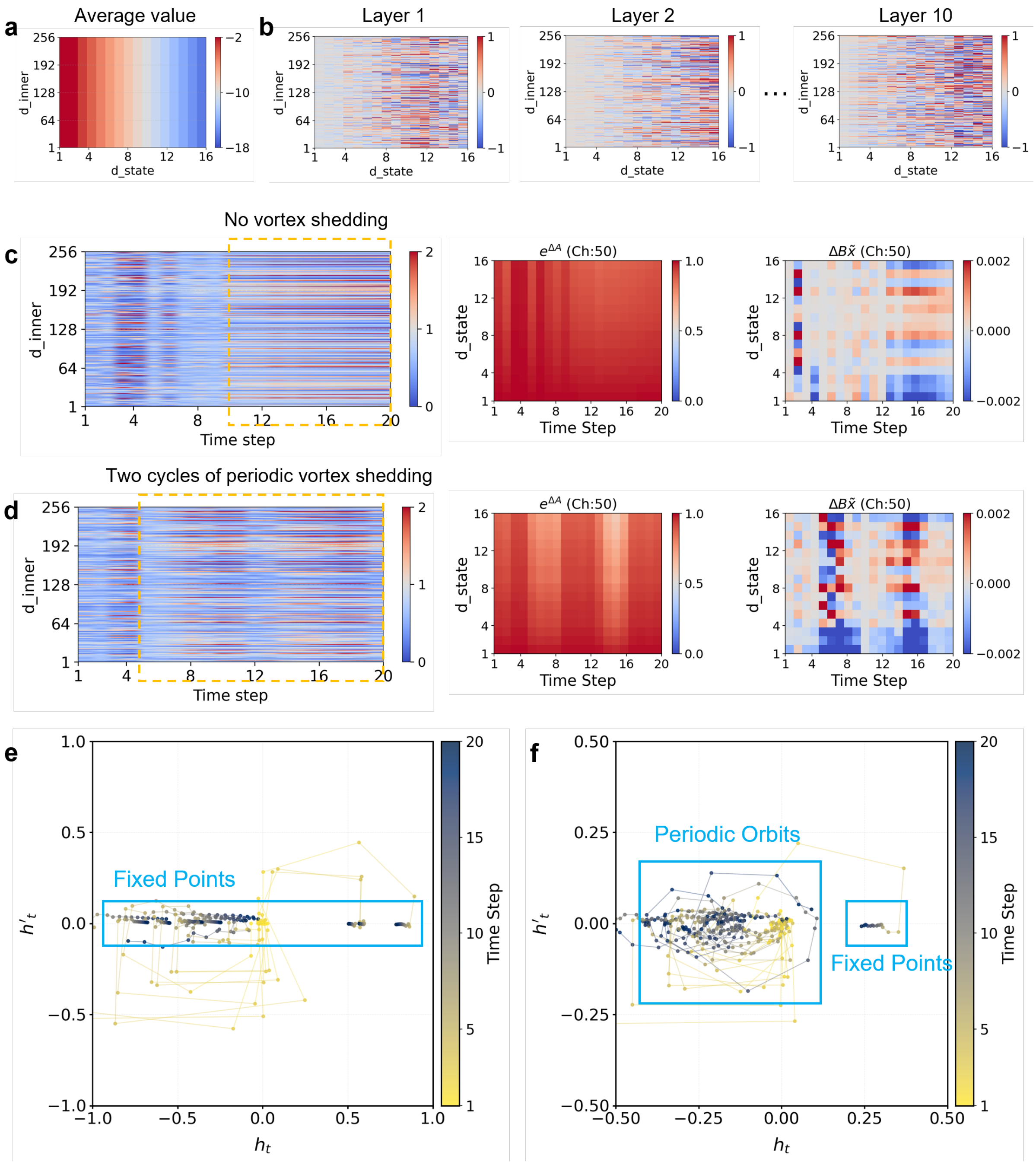

**Fig. 9.** Latent dynamics analysis in the Mamba block. **a** Average value of matrix $A$ across the 10-layer Mamba modules. **b** Matrix $A$ of each Mamba module layers after mean-centering. **c** State transition components $\Delta$, $e^{\Delta A}$ and $\Delta B\tilde{x}$ in the 10th Mamba layer under the non-vortex-shedding condition. **d** State transition components $\Delta$, $e^{\Delta A}$ and $\Delta B\tilde{x}$ in the 10th Mamba layer under the vortex shedding condition. **e** No vortex shedding: trajectories converge to fixed points in the $h_t$-$h_t^{'}$ phase space. **f** Vortex shedding:

trajectories evolve toward fixed points and periodic orbits in the $h_t$-$h_t^{'}$ phase space. d_inner and d_state are architectural hyperparameters in the Mamba module. Ch:50 denotes that the 50th channel in the d_inner-dimensional is chosen for visualization. $h_t^{'}$ is the discrete-time change of the hidden state $h_t$, i.e., $h_t^{'} = h_t - h_{t-1}$.

Periodic oscillations emerge only when the dissipation is embedded within the forcing term of $\Delta_t B_t \tilde{x}_t$. The latent state influences the transformed input $\tilde{x}_t$ through the autoregressive structure, and $\tilde{x}_t$ in turn determines the nonlinear feedback term $\Delta_t B_t \tilde{x}_t$. This recursive dependence closes the loop between state transition and input generation. During autonomous evolution, the nonlinear feedback injects structured energy into the latent state at specific phases of the trajectory. When this energy input balances the contraction induced by $e^{\Delta_t A}$ over one cycle, the system stabilizes onto a periodic orbit. Thus, periodic motion is not caused by instability of the linear operator but by a dynamic equilibrium between nonlinear feedback and linear dissipation within the closed-loop mapping. As shown in Fig. 9(d), stronger contraction in $e^{\Delta_t A}$ is accompanied by larger amplitudes of $\Delta_t B_t \tilde{x}_t$, forming a dynamic balance that sustains periodic oscillations. $\mathrm{d_inner}$ and $\mathrm{d_state}$ are architectural hyperparameters in the Mamba module. $\mathrm{Ch: 5}$ denotes that the 50th channel in the $\mathrm{d_inner}$-dimensional is chosen for visualization.

Based on this latent dynamic mechanism, the initial latent state $\boldsymbol{z}_0$ determines whether the trajectory converges toward a fixed-point attractor or evolves along a periodic orbit in $h_t$-$h_t'$ phase space, as shown in Fig. 9(e) and Fig 9(f). $h_t'$ is the discrete-time change of the hidden state $h_t$, i.e., $h_t' = h_t - h_{t-1}$. Even under flow conditions that support vortex shedding, each basin of attraction may contain both fixed points and periodic orbits. If $\boldsymbol{z}_0$ lies closer to a fixed point, the trajectory experiences a brief transient phase during which energy is dissipated by the linear transition operator $e^{\Delta_t A}$, ultimately converging to a steady value. If $\boldsymbol{z}_0$ is nearer to a periodic orbit, the trajectory is captured by the closed orbit, resulting in sustained oscillatory behaviour. This sensitive dependence on the initial latent state explains how different single starting points can lead to distinct macroscopic dynamical behaviours, encompassing both steady and oscillatory latent evolutions, depending on their position within the basin of attraction.

*E. Physics-Data Fusion Fine-tuning*

Once the temporal and spatial derivatives of each physical variable are obtained, the residuals of the governing physical equations at the query points, $\boldsymbol{x}_Q$, can be calculated. The resulting residuals are then fused with the original latent feature, yielding the latent fused feature that guides the Physics-Data fusion fine-tuning of the model predictions. The fine-tuning performance is quantitatively assessed using the Airfoil dataset.

The implementation involves a two-stage process: initial data-driven training across the entire airfoil dataset, followed by physics–data fusion constrained fine-tuning on specific samples. First, global prediction performance was evaluated using sparse training points. As shown in Fig. 10(a), random masking selected a subset of data points from the training set. At sampled locations, prediction accuracy remains comparable to full-data training, but errors increase significantly at unsampled points; for example, the global test error reaches 1.28 at a 10% sampling rate. Accuracy improves rapidly with more training points, yet in the airfoil dataset, gains become marginal beyond a 50% sampling rate due to the low-frequency and smooth nature of flow fields governed by the inviscid N-S equations. Beyond this point, additional data provide limited benefits. Incorporating Physics–Data fusion fine-tuning effectively leverages physical priors, enhancing predictive accuracy at unsampled or sparsely sampled locations and stabilising the learned flow field patterns.

Fig. 10(b) illustrates the evolution of global prediction error during fine-tuning in a selected case. The joint application of self-supervised loss and physics equation residual constraints causes a temporary rise in error followed by a steady decline, while physical residuals consistently decrease. Error reduction analysis reveals that prediction errors are consistently larger in the outer region (where points are sparser) when using randomly sampled training data. Notably, this region shows the most significant error reduction after incorporating physical constraints, demonstrating the effectiveness of fine-tuning in sparse data areas. Fig. 10(c) presents the average MSE reduction across 100 test cases, demonstrating improvements of 12.97% and 10.48% at sampling rates of 10% and 20%, respectively. $\mathcal{L}_{SP}$ is the MSE at sampled points. $\mathcal{L}_{AP}$ is the MSE at all points. $\mathcal{L}_{AP-FT}$ is the MSE at all points after fine-tuning. The results reveal an inverse relationship between sampling rate and error reduction magnitude. Cases with initially low MSE values exhibit more substantial reductions, reaching up to 25%, as shown in Fig. 10(d). In relative terms shown in Fig. 10(e), physical residuals can decrease by as much as 50% when initial predictions are already accurate. The blue dashed line in the figure serves as a two-part reference: it remains approximately flat in the low-MSE region, indicating a saturation effect where further reducing MSE does not yield significant additional gains in reduction rate. Beyond a certain MSE threshold, the line transitions to a slope of -1 in the log-log scale, indicating that the reduction rate approximately scales inversely with MSE. The bule dashed line symbolizes the ultimate capacity of our model. This trend suggests that samples with lower initial MSE tend to achieve greater relative improvement in physical consistency through fine-tuning. When ground truth values are used as constraints, error reductions up to 80% are achieved at a 10% sampling rate. However, such ground truth data are typically unavailable in a new test case. Under these conditions, the backbone network should first deliver sufficient baseline accuracy, after which physics-based fine-tuning can serve as an effective strategy for further performance enhancement.

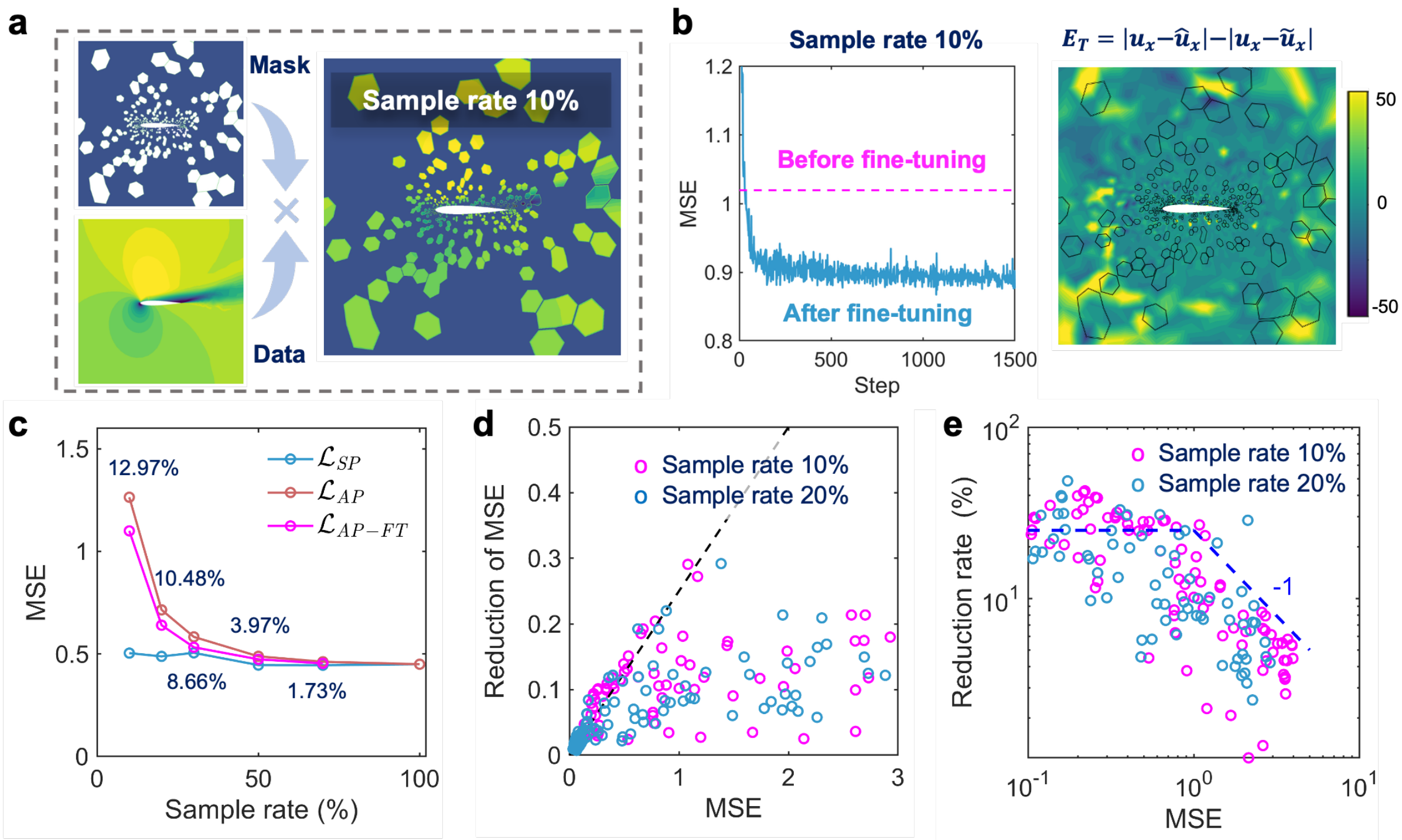

**Fig. 10.** Test results under sparse data training and Physics-Data fusion fine-tuning improvement effects. **a** Random masking sampling method for training data. **b** Ground truth error evolution during fine-tuning process. **c** MSE variation with respect to sampling rate of data. $\mathcal{L}_{SP}$ is the MSE at sampled points. $\mathcal{L}_{AP}$ is the MSE at all points. $\mathcal{L}_{AP-FT}$ is the MSE at all points after fine-tuning. **d** Distribution of MSE reduction values versus original MSE after fine-tuning. The dashed line has a slope of 0.25. **e** Distribution of post-fine-tuning MSE reduction ratios versus original MSE.

The spatiotemporal evolution of the physical field is controlled by the physical governing equations. Therefore, during model prediction and evaluation, it is not appropriate to solely use the MSE value. Instead, the evaluation should often incorporate the residuals from the physical governing equations. Fig. 11 presents the predicted results after fine-tuning using all data points. As shown in the figure, although the MSE values do not show significant variation after applying self-supervised constraints, the corresponding average residuals $\mathcal{R}$ from the physical equation values exhibit a notable order-of-magnitude reduction, indicating enhanced physical consistency across all samples. In the fine-tuning process, self-supervised constraints are applied to all points. All fine-tuning results are based on 200 steps of fine-tuning. Additionally, the two dashed lines in the figure represent the empirical relation $\mathcal{R} \propto 10^{2\cdot\mathrm{MSE}}$. As the MSE decreases below approximately 0.5, the value of $\mathcal{R}$ shows a rapid decline, approximately following an exponential relationship. However, once the MSE exceeds this threshold, the $\mathcal{R}$ values plateau and no longer decrease, suggesting a saturation effect where further decreases in MSE no longer significantly enhance physical consistency. A dual-metric evaluation system, MSE-$\mathcal{R}$, is better suited for assessing the quality of physical field generation. For new prediction tasks, where ground truth values may not be available and MSE cannot be computed, the $\mathcal{R}$ value then becomes the most important evaluation metric.

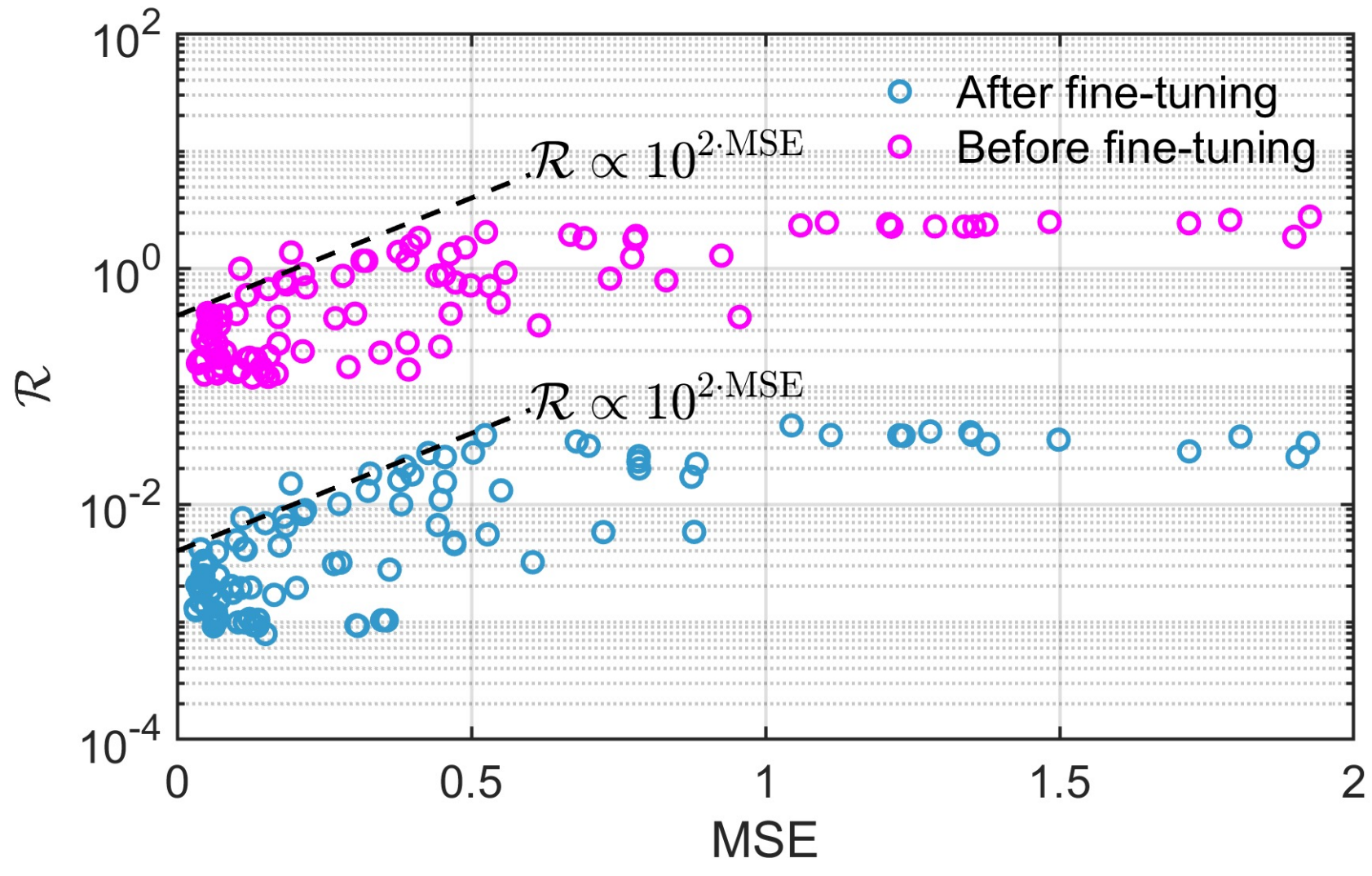

**Fig. 11.** MSE-$\mathcal{R}$ spatiotemporal physical field generation evaluation method.

During self-supervised fine-tuning in Fig. 11, the proportion of self-supervised points significantly influences the optimization of physics equation residuals. Specifically, a higher proportion of self-supervised points noticeably restricts the reduction of physics residuals, whereas a lower proportion enhances the model's ability to regulate the physical field. This regulatory mechanism enables the final physical field to achieve dual optimization: strictly satisfying physics-based constraints while preserving the key characteristics of the original flow field without significant alterations. Fig. 12 presents the fine-tuning results for an Airfoil case exhibiting vortex shedding. Specifically, Fig. 12(a) illustrates the loss curve during the fine-tuning process, where the backbone network is trained using the complete set of point data without sampling reduction. The light blue curve represents the loss between $\tilde{\phi}$ and $\hat{\phi}$, while $R_c$, $R_{mx}$, and $R_{my}$ denote the residuals of the continuity equation and the momentum equations in the x- and y-directions, respectively. After a certain training period, these residuals stabilize within the range of $10^{-5}$ to $10^{-3}$, indicating that the fine-tuning process effectively enforces physical conservation constraints. Fig. 12(b) further shows the variation of physical equation residuals with respect to $\xi_u$, where $\xi_u$ represents the proportion of data used for self-supervised training. A visual comparison of the residuals before and after fine-tuning is provided in Fig. 12(c) and Fig. 12(d). Specifically, Fig. 12(c) presents the momentum equation residuals, and Fig. 12(d) shows the residuals of the x-direction momentum equation, where the first row corresponds to the results before fine-tuning and the second row to those after fine-tuning. A notable reduction in residuals within the wake region can be clearly observed, further confirming the effectiveness of the fine-tuning process.

Fig. 12(e) highlights the error reduction in the wake region, where periodic vortex shedding occurs. The error metrics are defined as $E_{ux} = |u_x\text{-}\hat{u}_x|\text{-}|u_x\text{-}\tilde{u}_x|$ and $E_{uy} = |u_y\text{-}\hat{u}_y|\text{-}|u_y\text{-}\tilde{u}_y|$. The yellow region indicates areas where the fine-tuned model achieves improved accuracy under physical constraints. These results quantitatively demonstrate the enhanced predictive performance of the model in this critical flow region.

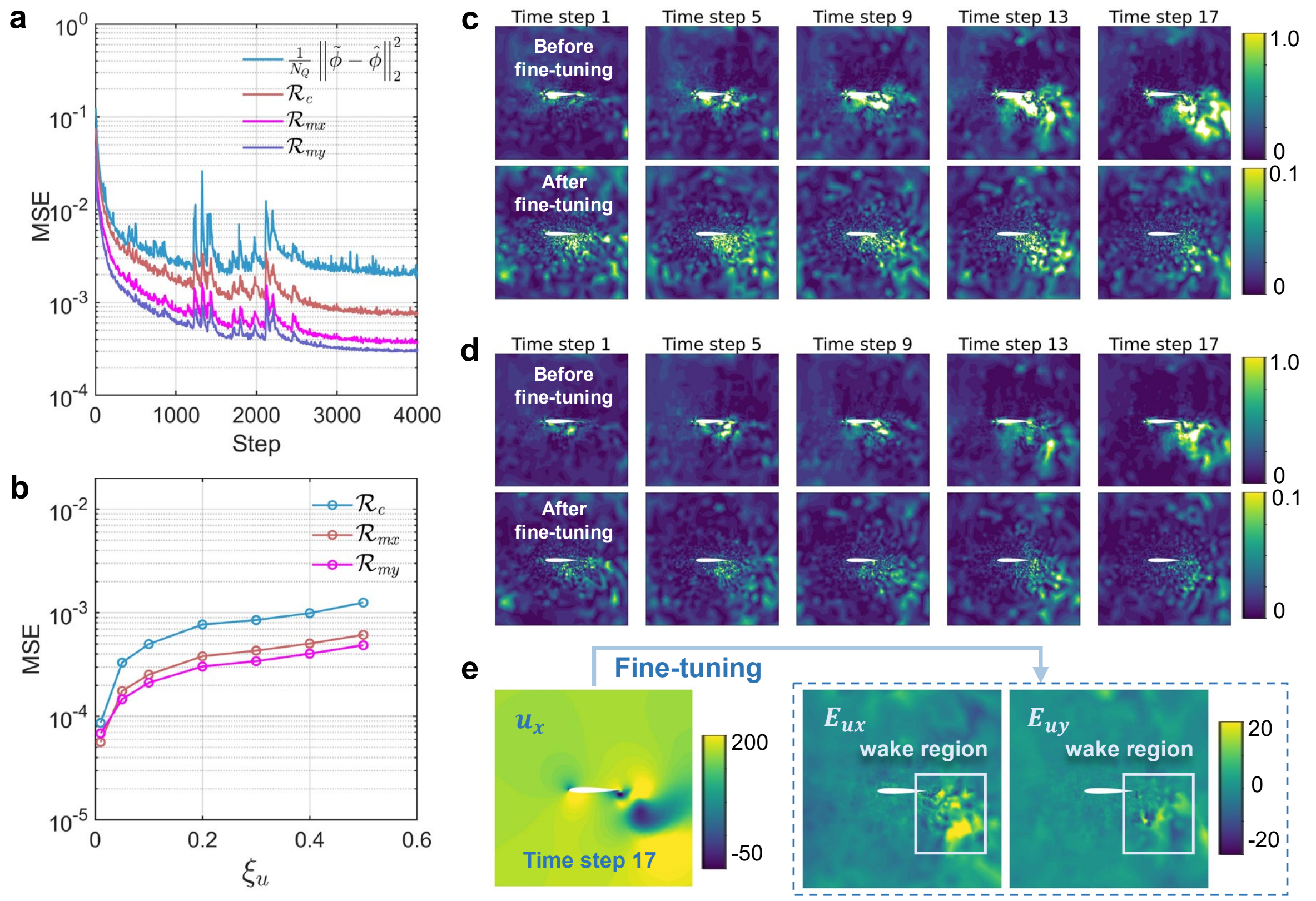

**Fig. 12.** Physical equation residuals results for an airfoil case after Fine-Tuning Process. **a** Loss curve during the fine-tuning process for the airfoil case. The backbone network is trained using the complete set of point data without any sampling reduction. Light blue represents the loss of $\tilde{\phi} - \hat{\phi}$. $\mathcal{R}_c$ is the continuity equation residuals, $\mathcal{R}_{mx}$ and $\mathcal{R}_{my}$ correspond to the momentum equation residuals in the $x$- and $y$-directions, respectively. **b** Physical equation residuals with respect to $\xi_u$, where $\xi_u$ is the propotion of the data used for self-supervision train. **c** Comparison of continuity equation residuals: the first row presents the momentum equation residuals before fine-tuning, while the second row illustrates the residuals after fine-tuning. **d** Comparison of momentum equation residuals in the x-direction: the first row displays the residuals before fine-tuning, and the second row shows the results after fine-tuning. **e** Error reduction in the wake region, in which $E_{ux} = |u_x\text{-}\hat{u}_x|\text{-}|u_x\text{-}\tilde{u}_x|$ and $E_{uy} = |u_y\text{-}\hat{u}_y|\text{-}|u_y\text{-}\tilde{u}_y|$. The yellow region represents the area where the fine-tuning results are improved under physical constraints.

## V. Conclusion

A spatiotemporal physical field generation model that integrates a physics–data fusion mechanism is proposed, realized via a Physics-Data fusion fine-tuning process. The backbone network utilizes a hybrid Mamba-Transformer architecture capable of processing unstructured point coordinates and physical field initialization information as inputs, while supporting arbitrary point queries for field information. To enhance the physical consistency of predictions, the model leverages physics–data fusion through fine-tuning with self-supervision. Through testing on diverse datasets and comparative evaluations with GEO-FNO, GINO, and TRANSOLVER models, the proposed model demonstrates superior predictive performance.

The latent space exhibits a hierarchical encode-decode mechanism, where shallow layers capture local temporal dependencies, intermediate layers encode robust representations, and the deepest layer decodes information relevant to the output. Latent dynamics balance linear dissipation and nonlinear feedback, leading trajectories to fixed points or periodic orbits depending on the initial state. This coupling explains how latent evolution governs the observed physical field patterns. By incorporating physics–data fusion during training, the model achieves approximately a 10% improvement in prediction accuracy when sparse training data are used. Additionally, this fusion process enhances prediction precision for highly transient and complex wake flow fields. A MSE-$\mathcal{R}$ evaluation method are suggested for assessing the accuracy and realism of physical field generation.

## Declaration of competing interest

The authors declare that they have no known competing finan-cial interests or personal relationships that could have appeared toinfluence the work reported in this paper.

## Data availability

Data will be made available on request.

## Acknowledgement

The authors greatly appreciate the financial support from the National Key R&D Program of China (2024YFF1500600)